\documentclass[sigconf]{acmart}
\AtBeginDocument{%
  \providecommand\BibTeX{{%
    \normalfont B\kern-0.5em{\scshape i\kern-0.25em b}\kern-0.8em\TeX}}}

\acmConference[DSHealth 2021]{2021 KDD Workshop on Applied Data Science for Healthcare}{August 14--18, 2021}{}
\begin{document}

\title{Midwifery Learning and Forecasting: Predicting Content Demand with User-Generated Logs}


%\email{webmaster@marysville-ohio.com}
%\affiliation{%
%  \institution{Institute for Clarity in Documentation}
%  \streetaddress{P.O. Box 1212}
%  \country{USA}

%\author{Lars Th{\o}rv{\"a}ld}
%\affiliation{%
%  \institution{The Th{\o}rv{\"a}ld Group}
%  \streetaddress{1 Th{\o}rv{\"a}ld Circle}
%\email{larst@affiliation.org}

\author{Anna Guitart}
\author{Ana Fern\'andez del R\'io}
\author{\'Africa Peri\'añez}
\email{africa@benshi.ai}
\affiliation{
  \institution{benshi.ai}
  \streetaddress{Passeig de Gracia, 74}
  \city{Barcelona}
  \state{}
  \postcode{08008}
  \country{Spain}}
  
\author{Lauren Bellhouse}
\email{lauren@maternity.dk}
\affiliation{
  \institution{Maternity Foundation}
  \streetaddress{Forbindelsesvej 3, 2. floor}
  \city{Copenhagen}
  \state{}
  \postcode{2100}
  \country{Denmark}}

\renewcommand{\shortauthors}{Guitart, et al.}

\begin{abstract}
  Every day, 800 women and 6,700 newborns die from complications related to pregnancy or childbirth. %, and there are 7,000 stillbirths. 
  A well-trained midwife can prevent most of these maternal and newborn deaths. Data science models together with logs generated by users of online learning applications for midwives can help to improve their learning competencies. The goal is to use these rich behavioral data to push digital learning towards personalized content and to provide an adaptive learning journey. In this work, we evaluate various forecasting methods to determine the interest of future users on the different kind of contents available in the app, broken down by profession and region.
\end{abstract}


%\ccsdesc[100]{Time series forecasting~deepAR}


\maketitle

\section{Introduction}

The rapid expansion of mobile health applications in low- and middle-income countries, and the large volume of data generated by their users, has created unprecedented opportunities for applying artificial intelligence (AI) to improve individual and population health~\cite{Hosny2019,Wahl2018}.
The application of data science models to the digital tools' behavioral logs of frontline healthcare workers and patients can lead to improvements in clinical research and practice, and health service delivery. And public health can use big datasets to promote healthy habits and ameliorate self-management, by providing people with health and well-being plans based on their particular medical and social circumstances~\cite{OConnor2018,Marsch2021}.

Every year 300,000 women and 5 million newborns die of causes related to pregnancy and childbirth~\cite{UNFPA2020, united2020levels}. Additionally, for every maternal death approximately 20 women suffer serious birth injuries~\cite{who2019trends}. Nearly all of these deaths and disabilities occur in low- and middle-income countries, and almost 90\% of them could be prevented if the woman gave birth with qualified assistance from a skilled midwife~\cite{Nove2021}. Additionally, 80\% of all newborn deaths result from conditions which are preventable and treatable, and for which proven, cost-efficient interventions exist~\cite{world2014every}. Almost all intrapartum and many antepartum stillbirths could be prevented with quality essential childbirth care and antenatal care~\cite{lawn2016stillbirths}.

Here we show an analysis of the logs from skilled birth attendants using the \emph{Safe Delivery App}~\cite{SDA}, a digital training and learning mobile application developed by the Maternity Foundation---a non-profit that develops digital learning tools to ensure all women and newborns have a safe childbirth~\cite{MF}. This work represents a step towards content personalization for midwives, in a sector that has traditionally been left out of big technological developments. Forecasting the demand for learning content by profession and region can lead to a better understanding of user habits and improve the management of campaigns~\cite{xu2019forecasting}. We apply several forecasting methods to evaluate their accuracy and production feasibility with the aim of using the outcomes for future experimentation and incentive analysis. Previous studies using similar methodologies and user logs can be found in~\cite{guitart2017forecasting,delrio2021}.

\begin{figure*}
  \includegraphics[width=\textwidth]{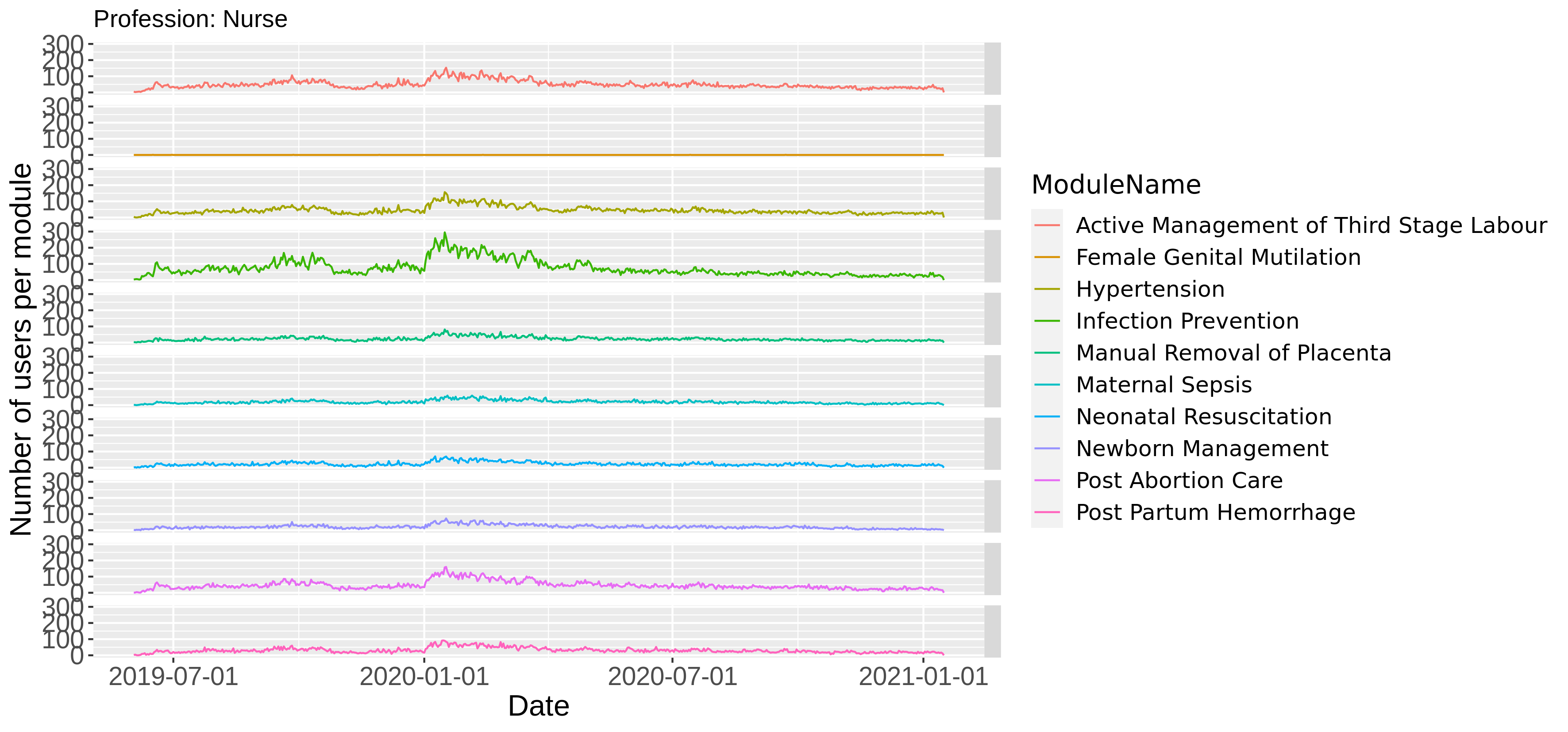}
  \caption{\emph{Nurses} module usage. Number of daily users who accessed a particular \emph{Safe Delivery App} module, for India \emph{nurses}.}
  \label{fig:nurse}
\end{figure*}

\subsection{Forecasting models}

We compare the performance of different time series forecasting methods in predicting the daily demand (in number of users) per type of content ({\it module}, in the app's language) and user's profession. Training and prediction were performed using the gluonTS~\cite{alexandrov2020gluonts}, keras~\cite{chollet2015keras} and mxnet~\cite{chen2015mxnet} Python libraries and the Forecast R package~\cite{Hyndman2008}. 

\subsubsection{Seasonal naïve forecaster}

The naïve forecast model~\cite{alexandrov2020gluonts} was used as a benchmark. Its forecasts are given by the exact values at the equivalent time points of the previous season. For prediction lengths larger than a season, the season is repeated multiple times, whereas for time series shorter than a season, the mean observed value is used as the prediction.

\subsubsection{Seasonal Autorregressive Integrated Moving Average (SARIMA) forecaster}

The SARIMA model~\cite{Box1976} was used as an additional benchmark, as it is one of the best performing and most widely used classical approaches to time series analysis and forecasting. At each time step, the time series value is a combination of regular and seasonal autoregressive (where the value depends on the previous values) and moving average (where the value depends on the previous errors) polynomials. In addition, one can take as many differences as needed in the original time series to make it stationary. 

\subsubsection{Neural networks with categorical embedding}

The last decade has seen the rapid growth of deep neural network architectures to tackle a great variety of problems~\cite{Lecun2015}, due to increasing computational resources and data availability, as well as improved methodology. One shortcoming of this approach is its difficulty to include categorical features, due to their lack of continuity. Entity embedding~\cite{guo2016entity} can be used to effectively learn the representation of categorical variables in multidimensional spaces, increasing their continuity and thus providing an intelligent way of using them as features in deep learning models. In particular, it overcomes the problems faced by the more traditional approach of one-hot encoding, namely the need for excessive computational power and its tendency to overfitting. 

\subsubsection{Autoregressive recurrent networks (DeepAR)}

The use of autoregressive recurrent networks to simultaneously predict many time series was introduced in~\cite{Salinas2020}. The method trains either long short-term memory (LSTM) or gated recurrent unit (GRU) networks, where the inputs at each time step  
%with target the time series value at that time step, and inputs 
are the covariates, the target value from the previous time step (which makes it autoregressive) and the previous output of the network (which makes it recurrent). A global model is learned from all the time series that can be used to generate probabilistic forecasts for the individual time series, each with its own individual distribution. This technique has been previously used in connection with healthcare in~\cite{papastefanopoulos2020covid}.

\subsubsection{Low-rank Gaussian copula processes}

This is a multivariate approach with deep learning elements, described as GP-Copula in~\cite{salinas2019high}. It combines a time series model based on autoregressive recurrent newtworks with a Gaussian copula process to parametrize the output distribution. This copula has low-rank structure in order to keep the number of parameters and computational complexity within reasonable bounds.

\begin{table*}
  \caption{Error metrics for the different forecasting models evaluated. The final errors are an average over the individual predictions using a rolling window from 2020-06-01 to 2020-12-01, with a forecasting horizon of 30 days.}
  \label{tab:error}
  \begin{tabular}{cccccc}
    \toprule
    Model & MASE & MAPE & sMAPE & RMSE & MSE \\
    \midrule
    \texttt{Seasonal Naïve} & 1.3221 & 0.7945 & 0.7294 & 8.2179 & 84.2975\\
    \texttt{ARIMA} & 0.8752 & 0.5569 & 0.7416 & 4.5020 & 21.2377\\
    %\texttt{Prophet} & 1.0061 & 0.6234 & 0.9558 & 5.5880 & 36.3663 \\
    \texttt{NN with categorical embeddings} & 1.0381 & 0.5283 & 0.6547 & 12.4945 & 250.0235 \\
    \texttt{DeepAR}  & 0.8110  & 0.4373 & 0.5689 & 4.4461 & 22.1728\\
    %\texttt{deepVAR} & 1.3564  & \textbf{0.6786} & 1.0817 & 10.1241 & 122.8886\\
    \texttt{GP-Copula} & 0.8209 & 0.4285 & 0.6564 & 4.4088 & 21.5326\\
    \bottomrule
  \end{tabular}
\end{table*}

\begin{figure*}
   \includegraphics[width=\textwidth]{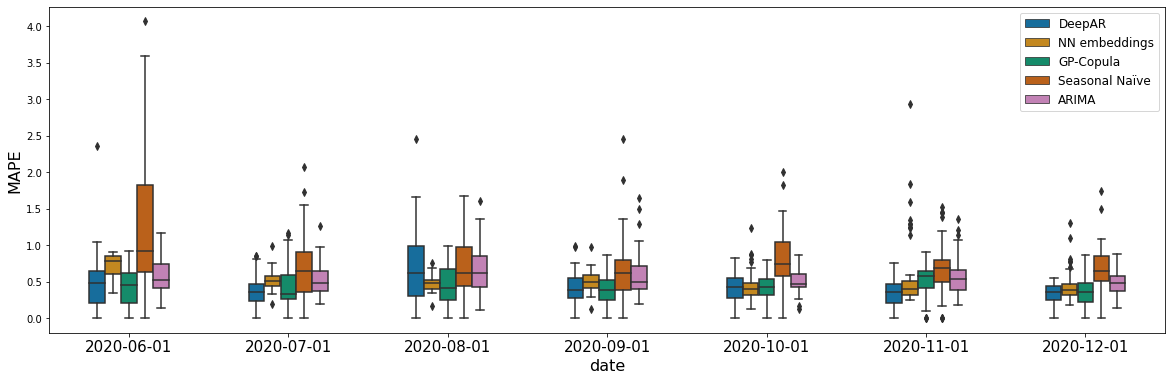}
   \caption{Mean Absolute Percentage Error (MAPE) boxplot distribution from individual time series displayed by model and by prediction month.}
   \label{fig:MAPE}
 \end{figure*}
 
\section{Modeling}

\subsection{Dataset}
Our dataset comprised user logs extracted from Maternity Foundation's \emph{Safe Delivery App}. This app targets skilled birth attendants around the world, empowering them to provide a safer birth
for mothers and newborns through evidence-based and up-to-date clinical guidelines on maternal and neonatal care, including the core components or "signal functions" of Basic Emergency Obstetric and Newborn Care. 

\emph{Safe Delivery App} is organized in several learning topics, known as modules: \emph{Active Management of Third Stage Labour}, \emph{Female Genital Mutilation}, \emph{Hypertension}, \emph{Infection Prevention}, \emph{Manual Removal of Placenta}, \emph{Maternal Sepsis}, \emph{Neonatal Resuscitation}, \emph{Newborn Management}, \emph{Post Abortion Care} and \emph{Post Partum Hemorrhage}. The possible professions of the app users (as self-reported by them on first login) are: \emph{Midwife}, \emph{Nurse}, \emph{Other skilled birth attendant}, \emph{Physician} or \emph{Student}.

Data were collected from 2019-05-01 to 2021-01-14, and the results shown correspond to a sample of 20,422 users from India. We built daily time series of module usage per profession, showing the number of professionals that accessed a particular module per day, and took them as a proxy for the demand for that specific content. Figure~\ref{fig:nurse} presents the time series for various modules in the case of~\emph{nurses}. Even though all modules show a similar overall usage trend, each series exhibits different scale and usage patterns. Similar series are obtained for the other professions. Our goal is to predict the app usage per profession, in order to personalize the content and get a better grasp of usage dynamics.

\subsection{Model specification}

The goal was to predict the daily values of the usage time series for each month and each module-profession combination, training the model with all the available data until the end of the previous month. Cross-validation was performed using a rolling window~\cite{racine2000consistent,cerqueira2020evaluating} from 2020-06-01 to 2020-12-01,considering all historical data before the prediction date for the training samples and 30 days of data before the prediction date for the test sample (training samples were split into training and validation sets). The final configuration was selected as the one that got the best results in the cross-validation rolling-window process.
All models used the profession and module as categorical features, and the day of the month, day of the week, month and year as covariates. The inclusion of COVID-19-related covariates and information on the \emph{Safe Delivery App} training for healthcare professionals was tested but did not result in any clear improvement.

Regarding the specifications of each method, the SARIMA forecasting was performed using the auto-ARIMA functionality, which means that all combinations of regular polynomials up to degree 5, seasonal polynomials up to degree 2, up to 2 regular and up to 1 seasonal differences were tried, using the Akaike Information Criterion to select the best of them.
Our neural network with categorical embeddings had 3 fully connected layers with 1000, 500 and 1 cells; the activation function for the dense layers was ReLU for the first and a sigmoid for the second layer, and we used the mean absolute error as the loss function and Adam as the optimization method. The best performing DeepAR model was found to be that using 20 2-layer LSTM cells, a negative binomial distribution, a dropout rate of 0.01, 300 epochs and a training batch size of 30. The selected GP-Copula variant had exactly the same settings, except that only 5 epochs were considered---as this method is much more computationally intensive and the model was already reaching convergence.

\subsection{Results}

Most of the forecasting models evaluated were able to capture the trend of the time series, with results differing mainly in the estimation of the daily patterns specific to each time series. Results are summarized in Table~\ref{tab:error}, which shows several error metrics averaged over across all monthly predictions. These are: Mean Absolute Scaled Error (MASE), Mean Absolute Percentage Error (MAPE), Symmetric Mean Absolute Percentage Error (sMAPE), Root Mean Squared Error (RMSE) and Mean Squared Error (MSE)~\cite{Botchkarev2019}. In Figure~\ref{fig:MAPE} the boxplot distribution of the individual MASE scores are displayed for each model and for each prediction month. We can observe that for more recent months, the scores tend to be lower, partially due to the enlargement of the historic data used in the training sample.

\begin{figure}
  \includegraphics[width=0.5\textwidth]{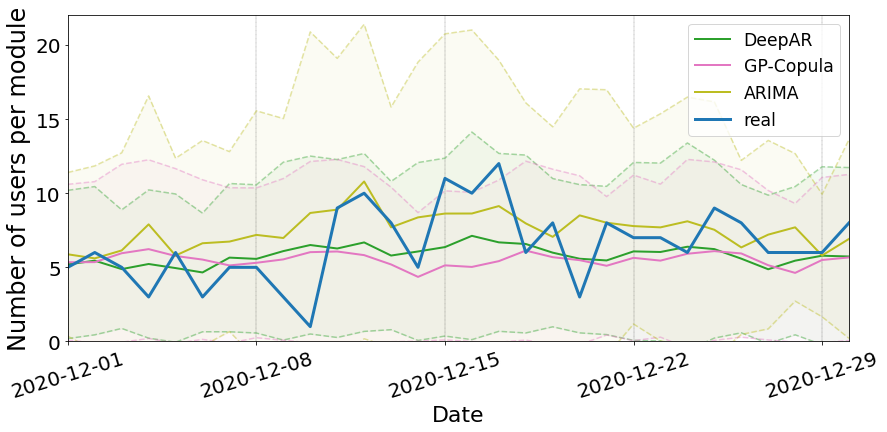}
  \includegraphics[width=0.5\textwidth]{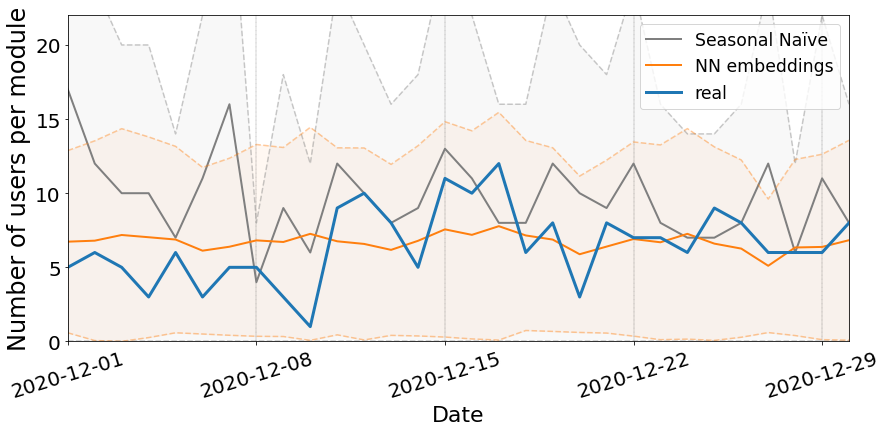}
  \caption{Real vs.\ forecasted values for the \emph{Infection Prevention} module demand by \emph{midwifes} on December 2020. The predictions and 50\% confidence intervals of the DeepAR, GP-Copula, ARIMA, NN with categorical embeddings and seasonal naïve models are shown.}
  %\Description{Forecast examples with the corresponding 50\% confidence interval for the time series corresponding to module \emph{Infection Prevention} and profession \emph{Midwive}.}
  \label{fig:forecast3}
\end{figure}

\begin{figure}
   \includegraphics[width=0.5\textwidth]{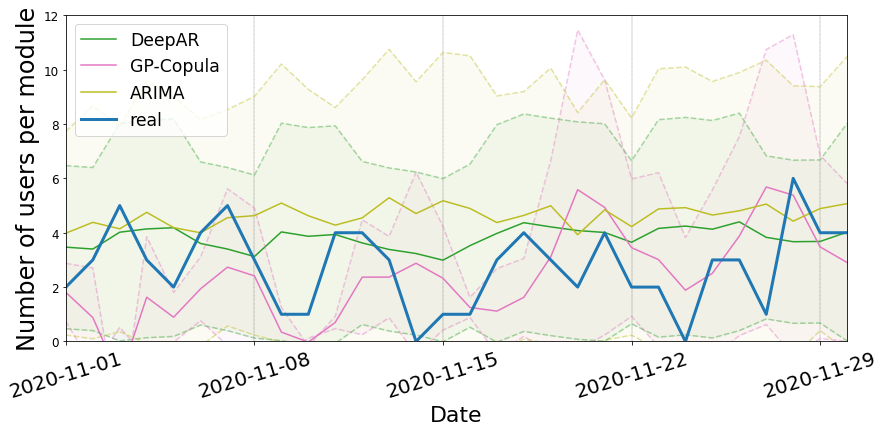}
   \includegraphics[width=0.5\textwidth]{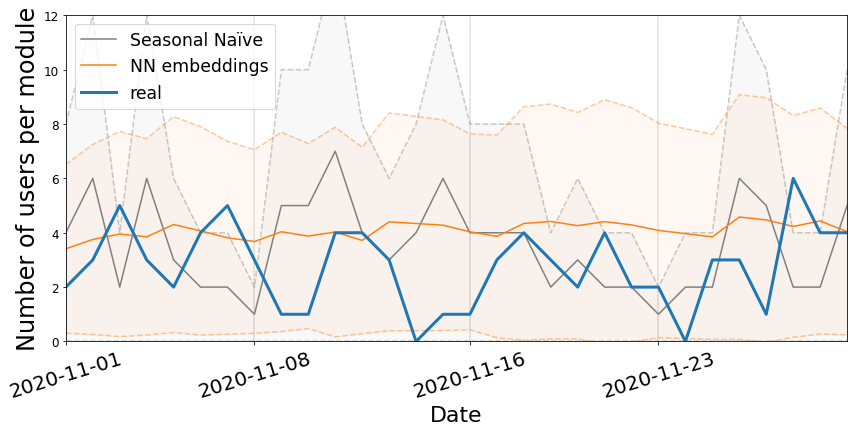}
   \caption{Real vs.\ forecasted values for the \emph{Active Management of Third Stage Labour} module demand by \emph{physicians} on November 2020. The predictions and 50\% confidence intervals of the DeepAR, GP-Copula, ARIMA, NN with categorical embeddings and seasonal naïve models are shown.}
  \label{fig:forecast5}
\end{figure}

Figures \ref{fig:forecast3} and \ref{fig:forecast5} shows an example of the forecasts for each model. It illustrates that, while the performance of the DeepAR and GP-Copula methods is relatively similar (left panel), the former shows a tighter 50\% confidence prediction interval that fits better the shape of the actual series. The ARIMA model also produced remarkably accurate predictions, which accounts for its extended use even nowadays that more sophisticated methods are available, although it shows a larger forecast uncertainty (as shown by the wider confidence interval). The forecasts for the other evaluated models are displayed in the right panel. The GP-Copula model trained over just 5 epochs performs similarly than the DeepAR model trained over 300 epochs, though it still needs more time and resources. For that reason, DeepAR would be the preferred option in a production environment. However, if higher accuracy were critical and there were no constraints on computational time and resources, the use of GP-Copula with an increased number of epochs would be justified.
%but they have obvious difficulties in accurately predicting both the level and shape of the observed time series.

\section{Summary and conclusion}

Overall, we found that the DeepAR and GP-Copula deep learning models are the most accurate for daily forecasting of the content demand. This result holds across different contents (modules) and user types (professions), as these two models show less error variability in the overall results of each individual time series. Other models such as Facebook's Prophet~\cite{taylor2018forecasting} and DeepVAR (the simplest multivariate extension of DeepAR)~\cite{salinas2019high}, both with their default settings, were also tested but performed worse than the ARIMA benchmark, so they were not included in the analysis. 

Although the evaluated dataset corresponds to India, this methodology shows potential to be applied to different countries or geographical areas, and also to additional contents. DeepAR constitutes a generalizable model that can correctly capture the trend behavior of the time series and anticipate user demand for a particular content depending on the user profile. We provided a solution that could be used in operational settings to get real-time demand estimates, due to the flexibility and speed of the model implementation.

\section*{Data}
All data used in this analysis comes from the Safe Delivery App logs and belongs to the Maternity Foundation. For inquiries regarding its use, please contact them at mail@maternity.dk.

\begin{acks}
The authors wish to thank Javier Grande and Wei Xiang Low for their careful review of the manuscript. This work was supported, in whole or in part, by the Bill \& Melinda Gates Foundation INV-022480. Under the grant conditions of the Foundation, a Creative Commons Attribution 4.0 Generic License has already been assigned to the Author Accepted Manuscript version that might arise from this submission.
\end{acks}

\bibliographystyle{ACM-Reference-Format}
\bibliography{main}


\begin{thebibliography}{29}

%%% the Web address.

\ifx \showCODEN    \undefined \def \showCODEN     #1{\unskip}     \fi
\ifx \showDOI      \undefined \def \showDOI       #1{#1}\fi
\ifx \showISBNx    \undefined \def \showISBNx     #1{\unskip}     \fi
\ifx \showISBNxiii \undefined \def \showISBNxiii  #1{\unskip}     \fi
\ifx \showISSN     \undefined \def \showISSN      #1{\unskip}     \fi
\ifx \showLCCN     \undefined \def \showLCCN      #1{\unskip}     \fi
\ifx \shownote     \undefined \def \shownote      #1{#1}          \fi
\ifx \showarticletitle \undefined \def \showarticletitle #1{#1}   \fi
\ifx \showURL      \undefined \def \showURL       {\relax}        \fi
% invisible to TeX
\providecommand\bibfield[2]{#2}
\providecommand\bibinfo[2]{#2}
\providecommand\natexlab[1]{#1}
\providecommand\showeprint[2][]{arXiv:#2}

\bibitem[\protect\citeauthoryear{Alexandrov, Benidis, Bohlke-Schneider,
  Flunkert, Gasthaus, Januschowski, Maddix, Rangapuram, Salinas, Schulz,
  et~al\mbox{.}}{Alexandrov et~al\mbox{.}}{2020}]%
        {alexandrov2020gluonts}
\bibfield{author}{\bibinfo{person}{Alexander Alexandrov},
  \bibinfo{person}{Konstantinos Benidis}, \bibinfo{person}{Michael
  Bohlke-Schneider}, \bibinfo{person}{Valentin Flunkert}, \bibinfo{person}{Jan
  Gasthaus}, \bibinfo{person}{Tim Januschowski}, \bibinfo{person}{Danielle~C
  Maddix}, \bibinfo{person}{Syama Rangapuram}, \bibinfo{person}{David Salinas},
  \bibinfo{person}{Jasper Schulz}, {et~al\mbox{.}}}
  \bibinfo{year}{2020}\natexlab{}.
\newblock \showarticletitle{GluonTS: Probabilistic and Neural Time Series
  Modeling in Python}.
\newblock \bibinfo{journal}{\emph{Journal of Machine Learning Research}}
  \bibinfo{volume}{21}, \bibinfo{number}{116} (\bibinfo{year}{2020}),
  \bibinfo{pages}{1--6}.
\newblock


\bibitem[\protect\citeauthoryear{Botchkarev}{Botchkarev}{2019}]%
        {Botchkarev2019}
\bibfield{author}{\bibinfo{person}{Alexei Botchkarev}.}
  \bibinfo{year}{2019}\natexlab{}.
\newblock \showarticletitle{A New Typology Design of Performance Metrics to
  Measure Errors in Machine Learning Regression Algorithms}.
\newblock \bibinfo{journal}{\emph{Interdisciplinary Journal of Information,
  Knowledge, and Management}}  \bibinfo{volume}{14} (\bibinfo{year}{2019}),
  \bibinfo{pages}{045–076}.
\newblock
\showISSN{1555-1237}
\urldef\tempurl%
\url{https://doi.org/10.28945/4184}
\showDOI{\tempurl}


\bibitem[\protect\citeauthoryear{Box and Jenkins}{Box and Jenkins}{1990}]%
        {Box1976}
\bibfield{author}{\bibinfo{person}{George Edward~Pelham Box} {and}
  \bibinfo{person}{Gwilym Jenkins}.} \bibinfo{year}{1990}\natexlab{}.
\newblock \bibinfo{booktitle}{\emph{Time Series Analysis, Forecasting and
  Control}}.
\newblock \bibinfo{publisher}{Holden-Day, Inc.}, \bibinfo{address}{USA}.
\newblock
\showISBNx{0816211043}


\bibitem[\protect\citeauthoryear{Cerqueira, Torgo, and Mozeti{\v{c}}}{Cerqueira
  et~al\mbox{.}}{2020}]%
        {cerqueira2020evaluating}
\bibfield{author}{\bibinfo{person}{Vitor Cerqueira}, \bibinfo{person}{Luis
  Torgo}, {and} \bibinfo{person}{Igor Mozeti{\v{c}}}.}
  \bibinfo{year}{2020}\natexlab{}.
\newblock \showarticletitle{Evaluating time series forecasting models: An
  empirical study on performance estimation methods}.
\newblock \bibinfo{journal}{\emph{Machine Learning}} \bibinfo{volume}{109},
  \bibinfo{number}{11} (\bibinfo{year}{2020}), \bibinfo{pages}{1997--2028}.
\newblock


\bibitem[\protect\citeauthoryear{Chen, Li, Li, Lin, Wang, Wang, Xiao, Xu,
  Zhang, and Zhang}{Chen et~al\mbox{.}}{2015}]%
        {chen2015mxnet}
\bibfield{author}{\bibinfo{person}{Tianqi Chen}, \bibinfo{person}{Mu Li},
  \bibinfo{person}{Yutian Li}, \bibinfo{person}{Min Lin},
  \bibinfo{person}{Naiyan Wang}, \bibinfo{person}{Minjie Wang},
  \bibinfo{person}{Tianjun Xiao}, \bibinfo{person}{Bing Xu},
  \bibinfo{person}{Chiyuan Zhang}, {and} \bibinfo{person}{Zheng Zhang}.}
  \bibinfo{year}{2015}\natexlab{}.
\newblock \bibinfo{title}{MXNet: A Flexible and Efficient Machine Learning
  Library for Heterogeneous Distributed Systems}.
\newblock
\newblock
\showeprint[arxiv]{1512.01274}
\urldef\tempurl%
\url{https://arxiv.org/abs/1512.01274}
\showURL{%
\tempurl}


\bibitem[\protect\citeauthoryear{Chollet et~al\mbox{.}}{Chollet
  et~al\mbox{.}}{2015}]%
        {chollet2015keras}
\bibfield{author}{\bibinfo{person}{Francois Chollet} {et~al\mbox{.}}}
  \bibinfo{year}{2015}\natexlab{}.
\newblock \bibinfo{booktitle}{\emph{Keras}}.
\newblock Keras.
\newblock
\urldef\tempurl%
\url{https://github.com/fchollet/keras}
\showURL{%
\tempurl}


\bibitem[\protect\citeauthoryear{del R\'io, Guitart, and Peri\'añez}{del R\'io
  et~al\mbox{.}}{2021}]%
        {delrio2021}
\bibfield{author}{\bibinfo{person}{Ana~Fern\'andez del R\'io},
  \bibinfo{person}{Anna Guitart}, {and} \bibinfo{person}{\'Africa
  Peri\'añez}.} \bibinfo{year}{2021}\natexlab{}.
\newblock \showarticletitle{{A Time Series Approach to Player Churn and
  Conversion in Videogames}}. In \bibinfo{booktitle}{\emph{{Intelligent Data
  Analysis, vol. 25, no. 1}}}. \bibinfo{pages}{177 -- 203}.
\newblock
\urldef\tempurl%
\url{https://doi.org/10.3233/IDA-194940}
\showDOI{\tempurl}


\bibitem[\protect\citeauthoryear{for Child Mortality~Estimation}{for Child
  Mortality~Estimation}{2020}]%
        {united2020levels}
\bibfield{author}{\bibinfo{person}{United Nations. Interagency~Group for Child
  Mortality~Estimation}.} \bibinfo{year}{2020}\natexlab{}.
\newblock \bibinfo{booktitle}{\emph{Levels \& Trends in Child Mortality: Report
  2020: Estimates Developed by the UN Inter-Agency Group for Child Mortality
  Estimation}}.
\newblock \bibinfo{publisher}{United Nations Children's Fund}.
\newblock


\bibitem[\protect\citeauthoryear{Foundation}{Foundation}{2021a}]%
        {MF}
\bibfield{author}{\bibinfo{person}{Maternity Foundation}.}
  \bibinfo{year}{2021}\natexlab{a}.
\newblock \bibinfo{title}{Maternity Foundation}.
\newblock \bibinfo{howpublished}{\url{https://www.maternity.dk/}}.
\newblock
\newblock
\shownote{Accessed: 2021-05-20.}


\bibitem[\protect\citeauthoryear{Foundation}{Foundation}{2021b}]%
        {SDA}
\bibfield{author}{\bibinfo{person}{Maternity Foundation}.}
  \bibinfo{year}{2021}\natexlab{b}.
\newblock \bibinfo{title}{Safe Delivery App}.
\newblock
  \bibinfo{howpublished}{\url{https://www.maternity.dk/safe-delivery-app/}}.
\newblock
\newblock
\shownote{Accessed: 2021-05-20.}


\bibitem[\protect\citeauthoryear{Fund}{Fund}{2020}]%
        {UNFPA2020}
\bibfield{author}{\bibinfo{person}{United Nations~Population Fund}.}
  \bibinfo{year}{2020}\natexlab{}.
\newblock \showarticletitle{Cost of Ending Preventable Maternal Deaths}. In
  \bibinfo{booktitle}{\emph{Costing the three transformative results}}.
  \bibinfo{pages}{11--17}.
\newblock
\urldef\tempurl%
\url{{https://www.unfpa.org/sites/default/files/pub-pdf/Transformative_results_journal_23-online.pdf}}
\showURL{%
\tempurl}


\bibitem[\protect\citeauthoryear{Guitart, Chen, Bertens, and
  Peri{\'a}{\~n}ez}{Guitart et~al\mbox{.}}{2018}]%
        {guitart2017forecasting}
\bibfield{author}{\bibinfo{person}{Anna Guitart}, \bibinfo{person}{Pei~Pei
  Chen}, \bibinfo{person}{Paul Bertens}, {and} \bibinfo{person}{{\'A}frica
  Peri{\'a}{\~n}ez}.} \bibinfo{year}{2018}\natexlab{}.
\newblock \showarticletitle{{F}orecasting {P}layer {B}ehavioral {D}ata and
  {S}imulating in-{G}ame {E}vents}. In \bibinfo{booktitle}{\emph{2018 IEEE
  Conference on Future of Information and Communication Conference (FICC)}}.
  \bibinfo{publisher}{IEEE}, \bibinfo{address}{Singapore}.
\newblock


\bibitem[\protect\citeauthoryear{Guo and Berkhahn}{Guo and Berkhahn}{2016}]%
        {guo2016entity}
\bibfield{author}{\bibinfo{person}{Cheng Guo} {and} \bibinfo{person}{Felix
  Berkhahn}.} \bibinfo{year}{2016}\natexlab{}.
\newblock \bibinfo{title}{Entity Embeddings of Categorical Variables}.
\newblock
\newblock
\showeprint[arxiv]{1604.06737}
\urldef\tempurl%
\url{https://arxiv.org/abs/1604.06737}
\showURL{%
\tempurl}


\bibitem[\protect\citeauthoryear{Hosny and Aerts}{Hosny and Aerts}{2019}]%
        {Hosny2019}
\bibfield{author}{\bibinfo{person}{Ahmed Hosny} {and} \bibinfo{person}{Hugo~JWL
  Aerts}.} \bibinfo{year}{2019}\natexlab{}.
\newblock \showarticletitle{Artificial intelligence for global health}.
\newblock \bibinfo{journal}{\emph{Science}} \bibinfo{volume}{366},
  \bibinfo{number}{6468} (\bibinfo{year}{2019}), \bibinfo{pages}{955--956}.
\newblock


\bibitem[\protect\citeauthoryear{Hyndman and Khandakar}{Hyndman and
  Khandakar}{2008}]%
        {Hyndman2008}
\bibfield{author}{\bibinfo{person}{R. Hyndman} {and} \bibinfo{person}{Yeasmin
  Khandakar}.} \bibinfo{year}{2008}\natexlab{}.
\newblock \showarticletitle{Automatic Time Series Forecasting: The forecast
  Package for R}.
\newblock \bibinfo{journal}{\emph{Journal of Statistical Software}}
  \bibinfo{volume}{27} (\bibinfo{year}{2008}), \bibinfo{pages}{1--22}.
\newblock


\bibitem[\protect\citeauthoryear{Lawn, Blencowe, Waiswa, Amouzou, Mathers,
  Hogan, Flenady, Fr{\o}en, Qureshi, Calderwood, et~al\mbox{.}}{Lawn
  et~al\mbox{.}}{2016}]%
        {lawn2016stillbirths}
\bibfield{author}{\bibinfo{person}{Joy~E Lawn}, \bibinfo{person}{Hannah
  Blencowe}, \bibinfo{person}{Peter Waiswa}, \bibinfo{person}{Agbessi Amouzou},
  \bibinfo{person}{Colin Mathers}, \bibinfo{person}{Dan Hogan},
  \bibinfo{person}{Vicki Flenady}, \bibinfo{person}{J~Frederik Fr{\o}en},
  \bibinfo{person}{Zeshan~U Qureshi}, \bibinfo{person}{Claire Calderwood},
  {et~al\mbox{.}}} \bibinfo{year}{2016}\natexlab{}.
\newblock \showarticletitle{Stillbirths: rates, risk factors, and acceleration
  towards 2030}.
\newblock \bibinfo{journal}{\emph{The Lancet}} \bibinfo{volume}{387},
  \bibinfo{number}{10018} (\bibinfo{year}{2016}), \bibinfo{pages}{587--603}.
\newblock


\bibitem[\protect\citeauthoryear{LeCun, Bengio, and Hinton}{LeCun
  et~al\mbox{.}}{2015}]%
        {Lecun2015}
\bibfield{author}{\bibinfo{person}{Yann LeCun}, \bibinfo{person}{Yoshua
  Bengio}, {and} \bibinfo{person}{Geoffrey Hinton}.}
  \bibinfo{year}{2015}\natexlab{}.
\newblock \showarticletitle{Deep Learning}.
\newblock \bibinfo{journal}{\emph{Nature}} \bibinfo{volume}{521},
  \bibinfo{number}{7553} (\bibinfo{year}{2015}), \bibinfo{pages}{436--444}.
\newblock
\urldef\tempurl%
\url{https://doi.org/10.1038/nature14539}
\showDOI{\tempurl}


\bibitem[\protect\citeauthoryear{Marsch}{Marsch}{2021}]%
        {Marsch2021}
\bibfield{author}{\bibinfo{person}{Lisa~A Marsch}.}
  \bibinfo{year}{2021}\natexlab{}.
\newblock \showarticletitle{Digital health data-driven approaches to understand
  human behavior}.
\newblock \bibinfo{journal}{\emph{Neuropsychopharmacology}}
  \bibinfo{volume}{46}, \bibinfo{number}{1} (\bibinfo{year}{2021}),
  \bibinfo{pages}{191--196}.
\newblock


\bibitem[\protect\citeauthoryear{Nove, Friberg, de~Bernis, McConville, Moran,
  Najjemba, ten Hoope-Bender, Tracy, and Homer}{Nove et~al\mbox{.}}{2021}]%
        {Nove2021}
\bibfield{author}{\bibinfo{person}{Andrea Nove}, \bibinfo{person}{Ingrid~K
  Friberg}, \bibinfo{person}{Luc de Bernis}, \bibinfo{person}{Fran McConville},
  \bibinfo{person}{Allisyn~C Moran}, \bibinfo{person}{Maria Najjemba},
  \bibinfo{person}{Petra ten Hoope-Bender}, \bibinfo{person}{Sally Tracy},
  {and} \bibinfo{person}{Caroline~SE Homer}.} \bibinfo{year}{2021}\natexlab{}.
\newblock \showarticletitle{Potential impact of midwives in preventing and
  reducing maternal and neonatal mortality and stillbirths: a Lives Saved Tool
  modelling study}.
\newblock \bibinfo{journal}{\emph{The Lancet Global Health}}
  \bibinfo{volume}{9}, \bibinfo{number}{1} (\bibinfo{year}{2021}),
  \bibinfo{pages}{e24--e32}.
\newblock


\bibitem[\protect\citeauthoryear{O'Connor}{O'Connor}{2018}]%
        {OConnor2018}
\bibfield{author}{\bibinfo{person}{Siobhan O'Connor}.}
  \bibinfo{year}{2018}\natexlab{}.
\newblock \showarticletitle{Big data and data science in health care: What
  nurses and midwives need to know}.
\newblock \bibinfo{journal}{\emph{Journal of Clinical Nursing}}
  \bibinfo{volume}{27}, \bibinfo{number}{15-16} (\bibinfo{year}{2018}),
  \bibinfo{pages}{2921--2922}.
\newblock
\urldef\tempurl%
\url{https://doi.org/10.1111/jocn.14164}
\showDOI{\tempurl}


\bibitem[\protect\citeauthoryear{Organization and UNICEF}{Organization and
  UNICEF}{2014}]%
        {world2014every}
\bibfield{author}{\bibinfo{person}{World~Health Organization} {and}
  \bibinfo{person}{UNICEF}.} \bibinfo{year}{2014}\natexlab{}.
\newblock \showarticletitle{Every newborn: an action plan to end preventable
  deaths}.
\newblock  (\bibinfo{year}{2014}).
\newblock


\bibitem[\protect\citeauthoryear{Papastefanopoulos, Linardatos, and
  Kotsiantis}{Papastefanopoulos et~al\mbox{.}}{2020}]%
        {papastefanopoulos2020covid}
\bibfield{author}{\bibinfo{person}{Vasilis Papastefanopoulos},
  \bibinfo{person}{Pantelis Linardatos}, {and} \bibinfo{person}{Sotiris
  Kotsiantis}.} \bibinfo{year}{2020}\natexlab{}.
\newblock \showarticletitle{Covid-19: A comparison of time series methods to
  forecast percentage of active cases per population}.
\newblock \bibinfo{journal}{\emph{Applied Sciences}} \bibinfo{volume}{10},
  \bibinfo{number}{11} (\bibinfo{year}{2020}), \bibinfo{pages}{3880}.
\newblock


\bibitem[\protect\citeauthoryear{Racine}{Racine}{2000}]%
        {racine2000consistent}
\bibfield{author}{\bibinfo{person}{Jeff Racine}.}
  \bibinfo{year}{2000}\natexlab{}.
\newblock \showarticletitle{Consistent cross-validatory model-selection for
  dependent data: hv-block cross-validation}.
\newblock \bibinfo{journal}{\emph{Journal of econometrics}}
  \bibinfo{volume}{99}, \bibinfo{number}{1} (\bibinfo{year}{2000}),
  \bibinfo{pages}{39--61}.
\newblock


\bibitem[\protect\citeauthoryear{Salinas, Bohlke-Schneider, Callot, Medico, and
  Gasthaus}{Salinas et~al\mbox{.}}{2019}]%
        {salinas2019high}
\bibfield{author}{\bibinfo{person}{David Salinas}, \bibinfo{person}{Michael
  Bohlke-Schneider}, \bibinfo{person}{Laurent Callot}, \bibinfo{person}{Roberto
  Medico}, {and} \bibinfo{person}{Jan Gasthaus}.}
  \bibinfo{year}{2019}\natexlab{}.
\newblock \showarticletitle{High-dimensional multivariate forecasting with
  low-rank Gaussian Copula Processes}. In \bibinfo{booktitle}{\emph{Advances in
  Neural Information Processing Systems}},
  \bibfield{editor}{\bibinfo{person}{H.~Wallach},
  \bibinfo{person}{H.~Larochelle}, \bibinfo{person}{A.~Beygelzimer},
  \bibinfo{person}{F.~d\textquotesingle Alch\'{e}-Buc},
  \bibinfo{person}{E.~Fox}, {and} \bibinfo{person}{R.~Garnett}} (Eds.),
  Vol.~\bibinfo{volume}{32}. \bibinfo{publisher}{Curran Associates, Inc.},
  \bibinfo{address}{Vancouver, Canada}.
\newblock
\urldef\tempurl%
\url{https://proceedings.neurips.cc/paper/2019/file/0b105cf1504c4e241fcc6d519ea962fb-Paper.pdf}
\showURL{%
\tempurl}


\bibitem[\protect\citeauthoryear{Salinas, Flunkert, Gasthaus, and
  Januschowski}{Salinas et~al\mbox{.}}{2020}]%
        {Salinas2020}
\bibfield{author}{\bibinfo{person}{David Salinas}, \bibinfo{person}{Valentin
  Flunkert}, \bibinfo{person}{Jan Gasthaus}, {and} \bibinfo{person}{Tim
  Januschowski}.} \bibinfo{year}{2020}\natexlab{}.
\newblock \showarticletitle{DeepAR: Probabilistic forecasting with
  autoregressive recurrent networks}.
\newblock \bibinfo{journal}{\emph{International Journal of Forecasting}}
  \bibinfo{volume}{36}, \bibinfo{number}{3} (\bibinfo{year}{2020}),
  \bibinfo{pages}{1181--1191}.
\newblock
\showISSN{0169-2070}
\urldef\tempurl%
\url{https://doi.org/10.1016/j.ijforecast.2019.07.001}
\showDOI{\tempurl}


\bibitem[\protect\citeauthoryear{Taylor and Letham}{Taylor and Letham}{2018}]%
        {taylor2018forecasting}
\bibfield{author}{\bibinfo{person}{Sean~J Taylor} {and}
  \bibinfo{person}{Benjamin Letham}.} \bibinfo{year}{2018}\natexlab{}.
\newblock \showarticletitle{Forecasting at scale}.
\newblock \bibinfo{journal}{\emph{The American Statistician}}
  \bibinfo{volume}{72}, \bibinfo{number}{1} (\bibinfo{year}{2018}),
  \bibinfo{pages}{37--45}.
\newblock


\bibitem[\protect\citeauthoryear{Wahl, Cossy-Gantner, Germann, and
  Schwalbe}{Wahl et~al\mbox{.}}{2018}]%
        {Wahl2018}
\bibfield{author}{\bibinfo{person}{Brian Wahl}, \bibinfo{person}{Aline
  Cossy-Gantner}, \bibinfo{person}{Stefan Germann}, {and}
  \bibinfo{person}{Nina~R Schwalbe}.} \bibinfo{year}{2018}\natexlab{}.
\newblock \showarticletitle{Artificial intelligence (AI) and global health: how
  can AI contribute to health in resource-poor settings{?}}
\newblock \bibinfo{journal}{\emph{BMJ global health}} \bibinfo{volume}{3},
  \bibinfo{number}{4} (\bibinfo{year}{2018}), \bibinfo{pages}{e000798}.
\newblock


\bibitem[\protect\citeauthoryear{World Health~Organization and Bank}{World
  Health~Organization and Bank}{2019}]%
        {who2019trends}
\bibfield{author}{\bibinfo{person}{United Nations Population~Fund World
  Health~Organization, UNICEF} {and} \bibinfo{person}{The~World Bank}.}
  \bibinfo{year}{2019}\natexlab{}.
\newblock \bibinfo{title}{Trends in maternal mortality: 2000 to 2017: estimates
  by WHO, UNICEF}.
\newblock
\newblock


\bibitem[\protect\citeauthoryear{Xu and Chan}{Xu and Chan}{2019}]%
        {xu2019forecasting}
\bibfield{author}{\bibinfo{person}{Shuojiang Xu} {and}
  \bibinfo{person}{Hing~Kai Chan}.} \bibinfo{year}{2019}\natexlab{}.
\newblock \showarticletitle{Forecasting medical device demand with online
  search queries: A big data and machine learning approach}.
\newblock \bibinfo{journal}{\emph{Procedia Manufacturing}}
  \bibinfo{volume}{39} (\bibinfo{year}{2019}), \bibinfo{pages}{32--39}.
\newblock


\end{thebibliography}

\end{document}